\documentclass{article}
\usepackage{spconf,amsmath,graphicx}
\usepackage{amsmath,graphicx}
\usepackage{amsfonts}
\usepackage{color}
\usepackage{caption}
\usepackage{subfig}

\include{definitions}
\title{Recovery of point clouds on surfaces: application to image reconstruction}
\name{Sunrita Poddar, Mathews Jacob\thanks{This  work  is  supported  by NIH 1R01EB019961-01A1 and ONR-N000141310202.}}
\address{Department of Electrical and Computer Engineering, University of Iowa, IA, USA}
\begin{document}
%
\maketitle
\vspace{-2em}
\begin{abstract}
We introduce a framework for the recovery of points on a smooth surface in high-dimensional space, with application to dynamic imaging. We assume the surface to be the zero-level set of a bandlimited function. We show that the exponential maps of the points on the surface satisfy annihilation relations, implying that they lie in a finite dimensional subspace. We rely on nuclear norm minimization of the maps to recover the points from noisy and undersampled measurements. Since this direct approach suffers from the curse of dimensionality, we introduce an iterative reweighted algorithm that uses the "kernel trick". The resulting algorithm has similarities to iterative algorithms used in graph signal processing (GSP); this framework can be seen as a continuous domain alternative to discrete GSP theory. The use of the algorithm in recovering free breathing and ungated cardiac data shows the potential of this framework in practical applications.
\end{abstract}
\vspace{-1em}
\section{Introduction}
\vspace{-1em}
The recovery of signals that lie on a manifold/surface has received extensive attention in the recent years. These methods model the high-dimensional data as points localized to low-dimensional manifolds.  For example, patch-based image processing methods such as BM3D implicitly use the structure of the patch manifold \cite{bm3d, yasir, mohsin2015iterative}, while we \cite{storm, poddar2014joint} and others \cite{nakarmi, rueckert} have recently used the manifold structure of images in a dynamic time series. Manifold embedding methods are also widely used in machine learning as visualization tools. 

The main focus of this paper is to introduce a continuous domain perspective for denoising/regularization of points (e.g patches or images), which are assumed to be drawn from a smooth surface in a very high dimensional space. This surface can be represented as the zero-level set of a bandlimited potential function. We note that the level-set model can account for a large variety of surfaces and is widely used in image processing. We observe that the potential function is zero for any point on the surface, which we term as an annihilation relation. We show that the annihilation relation can be expressed as a weighted linear combination of exponential feature maps of the point. The dimension of the feature map is equal to the bandwidth of the potential function. When the bandwidth is overestimated, there are multiple such annihilation relations, suggesting that the exponential feature maps of the points on the surface lie in a finite dimensional space. We show that the finite dimensional nature of the maps translates to a low-rank kernel matrix, computed from the points using a shift invariant kernel function such as the Dirichlet and Gaussian kernels. This demonstrates the direct link between superresolution signal recovery \cite{gregpapers} and kernel methods, which are widely used in machine learning.

We propose to use the nuclear norm of the exponential feature maps of the points as a regularizer in inverse problems, including denoising and recovery from incomplete data. We use the "kernel trick" to avoid the explicit computation of the maps or the surface, thus keeping the computational complexity manageable. We rely on an iterative reweighted algorithm to recover the denoised points. The resulting algorithm has similarities to iterative non-local methods \cite{wendy,graph} that are widely used in image processing and graph signal processing. Specifically, it alternates between the estimation of a graph Laplacian, which specifies the connectivity of the points, and the smoothing of points guided by the graph Laplacian. The proposed framework can be thought of as a more principled alternative to the above heuristic approaches. We demonstrate the utility of the algorithm in the challenging application involving the recovery of free breathing and ungated cardiac MRI data from highly undersampled measurements. The experiments show the great potential of this method in imaging applications, where patches/images can be assumed to lie on a manifold or smooth surface in high-dimensional space.

This work is built upon our prior work \cite{gregpapers} and the recent work by Ongie et al., which considered polynomial kernels \cite{gregvariety}. The direct extension of \cite{gregpapers} to recover the patch/image surface is computationally prohibitive due to the curse of dimensionality. The main focus of this work is to generalize \cite{gregvariety} to shift invariant kernels, which are more widely used in practice. Our approach uses an implicit representation of the surface using bandlimited or Gaussian functions, which is more useful in practical applications. In addition, our work shows the connections with graph Laplacian based methods that are widely used in graph signal processing as well as patch-based methods. The application to MRI has similarities to \cite{storm, nakarmi}, which estimate the kernel subspace structure from navigator data. In contrast, our focus is to develop a navigator-free reconstruction strategy. Besides, \cite{nakarmi} relies on an explicit mapping between the data and higher dimensional maps, whereas our scheme only works in the original domain. 
\vspace{-2.5em}
\section{Bandlimited surfaces \& annihilation}
\vspace{-1em}
We assume the point cloud to be supported on a surface in $\left[-1/2,1/2\right]^n$, which is the zero-level set of a bandlimited potential function:
\begin{equation}
\label{implicit}
\{\mathbf r \in \mathbb R^n|\psi(\mathbf r)=0\} ~\mbox{where}~\psi(\mathbf r) = \sum_{\mathbf k \in \Lambda} \mathbf c_{\mathbf k} e^{j~2\pi \mathbf k^T \mathbf r}
\end{equation}
Here, $\mathbf c_{\mathbf k}; \mathbf k\in \Lambda$ is the smallest set of coefficients (minimal set) that satisfies the above relation. $\Lambda\subset \mathbb Z^{n}$ is a set of contiguous locations that indicates the support of the Fourier series coefficients of $\psi$. Note that the above level set representation is widely used in image processing applications and can represent a large class of open and closed shapes \cite{gregpapers}.

Consider an arbitrary point $\mathbf x$ on the above surface \eqref{implicit}. By definition \eqref{implicit}, we have the annihilation relation $\psi(\mathbf x) = \sum_{\mathbf k \in \Lambda} \mathbf c_{\mathbf k} e^{j~2\pi\mathbf k^T \mathbf x} = 0$. The annihilation relation can be re-expressed in terms of the non-linear feature map $\phi_{\Lambda}(\mathbf x)$ as $
\mathbf c^T\phi_{\Lambda}(\mathbf x) = 0$, where $\phi: \mathbb R^n \rightarrow \mathbb C^{|\Lambda|}$ is defined as:
\begin{equation}
\phi_{\Lambda}(\mathbf x) = \begin{bmatrix} e^{j2\pi\mathbf k_1^T\mathbf x}&
 \ldots&
  e^{j2\pi\mathbf k_{|\Lambda|}^T\mathbf x}
\end{bmatrix}^T
\end{equation}

We now consider a collection of $N > |\Lambda|$ points on the surface, stacked into a matrix $
\mathbf X = \left[\mathbf x_1,\mathbf x_2,\ldots \mathbf x_N\right]$. With the same argument as above, we have: 
\begin{equation}
\label{matrixannihilation}
\mathbf c^T \underbrace{\begin{bmatrix}
\phi_{\Lambda}(\mathbf x_1),\ldots \phi_{\Lambda}(\mathbf x_N)
\end{bmatrix}
}_{\Phi_{\Lambda}(\mathbf X)} = 0,
\end{equation}
where $\Phi_{\Lambda}(\mathbf X) \in \mathbb C^{|\Lambda| \times N}$. The relation \eqref{matrixannihilation} implies that there is one vector in the null space of the feature matrix $\Phi_{\Lambda}(\mathbf X)$. Since $\mathbf c$ is the unique minimal filter, there is no other null space vector that satisfies the above relation. This implies that the rank of the feature matrix is $|\Lambda|-1$. In practice, we do not know the exact Fourier support of $\psi$; we can overestimate the support to $\Gamma \supseteq \Lambda $. If we considered a mapping $\phi_{\Gamma}: \mathbb R^n \rightarrow \mathbb C^{|\Gamma|}$, any function $\psi'$ that is bandlimited to $\Gamma$ is of the form $\psi'(\mathbf r) = \psi(\mathbf r)\eta(\mathbf r)$, where $\eta(\mathbf r)$ is any function. Then, for any $\mathbf r$ such that $\psi(\mathbf r)=0$, we also have $\psi'(\mathbf r)=0$. One can find $\Gamma:\Lambda$ linearly independent $\eta$ functions, where $\Gamma:\Lambda$ denotes the set of translations of $\Lambda$ in $\Gamma$ \cite{gregpapers}. We have:
\begin{equation}
\mbox{rank}\left(\Phi_{\Gamma}(\mathbf X)\right) \leq |\Gamma|-|\Gamma:\Lambda|
\end{equation}
provided $\Lambda \subseteq \Gamma$ and the points $\{\mathbf x_i\}$ are on the surface \eqref{implicit}.

\vspace{-1em}
\subsection{Surface recovery from noisy point cloud}
\vspace{-0.5em}
The least square estimation of the coefficients from the data points can be posed as a minimization of the criterion $\mathcal C(\mathbf c) = \sum_{i=1}^N \|\psi(\mathbf x_i)\|^2= \mathbf c^T \mathbf Q_{\Gamma} \mathbf c$, where $\mathbf Q_{\Gamma}=\sum_{i=1}^N \phi_{\Gamma}(\mathbf x_i)\phi_{\Gamma}(\mathbf x_i)^T$. The estimation of the coefficients can be posed as the eigenvalue problem:
\begin{equation}
\label{eigen}
\mathbf c^* = \arg \min_{\mathbf c} \mathbf c^T ~\mathbf Q_{\Gamma}~ \mathbf c ~~\mbox{such that }~~ \|\mathbf c\|^2 = 1,
\end{equation}
whose solution is the minimum eigenvector  of the matrix $\mathbf Q_{\Gamma}$.

\begin{figure}[t!]
\centering
\includegraphics[width=0.3\textwidth]{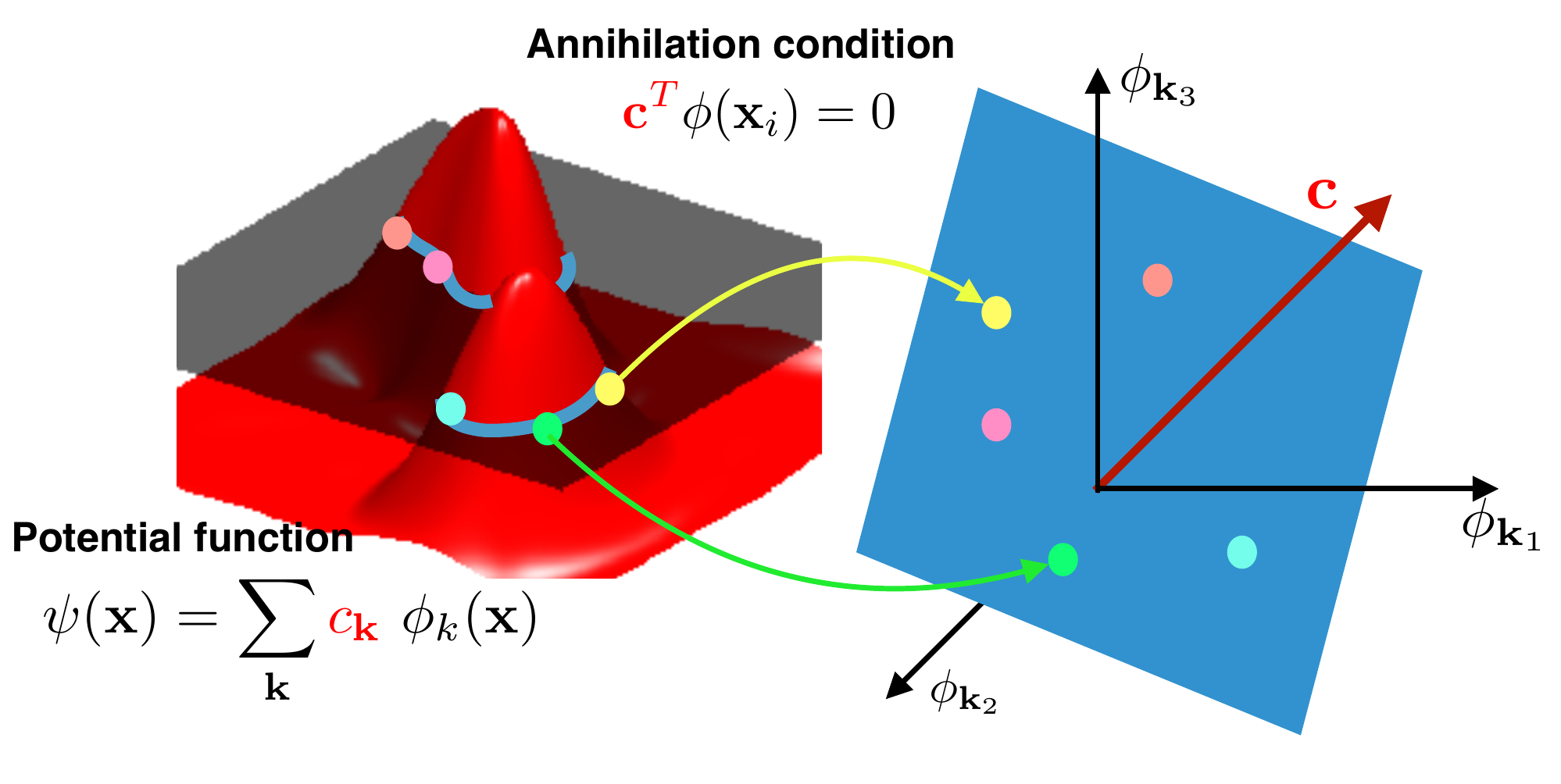}
\vspace{-1.4em}
\label{illus}
\caption{Illustration of the annihilation relations in 2-D. We assume that the curves/surface is the zero-level set of the potential function $\psi(\mathbf x)$. Each point on the curve/surface satisfies $\psi(\mathbf x_i)=0=\mathbf c^T\phi(\mathbf x_i)$, which can be seen as an annihilation relation in the non-linear feature space $\phi(\mathbf x)$. Specifically, the maps of the points lie on a plane orthogonal to $\mathbf c$.}\vspace{-1.5em}
\end{figure}

Note that bandlimited $\psi$ functions are often oscillatory, resulting in spurious zero locations. One may instead use weighted basis functions $\phi'(\mathbf x) = \mathbf D ~\phi(\mathbf x)$ to obtain smoother curves. We choose the weight matrix as a diagonal matrix with diagonal entries $e^{-\pi^2 \sigma^2\|\mathbf k\|^2}$.

\vspace{-1em}
\subsection{Relation to shift invariant non-linear kernels}
\vspace{-0.5em}
The above explicit approach is feasible when the dimension of the space $n$ is small. However, the dimension of the feature space grows as a power of $n$, making this approach impractical in applications involving clouds of images or patches. Hence, we rely on the right null space relations. Since the rank of the feature matrix $\Phi_{\Gamma}$ is $r\leq |\Gamma|-|\Gamma:\Lambda|$, we can find $N-r$ vectors $\mathbf v_i$ such that $\Phi_{\Gamma}(\mathbf X)~\mathbf v_i = \mathbf 0$, or equivalently, 
\begin{equation}
\underbrace{\Phi_{\Gamma}(\mathbf X)^H\Phi_{\Gamma}(\mathbf X)}_{\mathbf K^{\Gamma}}\mathbf v_i = \mathbf 0,
\end{equation}
The entries of the $N\times N$ Gram matrix $\mathbf K^{\Gamma}$ are: 
\begin{equation}
\mathbf K_{i,j}^{\Gamma} = \phi_{\Gamma}(\mathbf x_i)^H\phi_{\Gamma}(\mathbf x_j) =\underbrace{ \sum_{k\in \Gamma}  e^{\left(j~2\pi\mathbf k^T \left(\mathbf x_j-\mathbf x_i\right)\right)}}_{\kappa_{\Gamma}(\mathbf x_j-\mathbf x_i)},
\end{equation}
where $\kappa_{\Gamma}(\mathbf r)$ is shift invariant. When $\Gamma$ is a centered cube in $\mathbf R^n$, we have $\kappa_{\Gamma}(\mathbf r) = {\rm D}_{\Gamma}(\mathbf r)$, where $D_{\Gamma}$ is the Dirichlet kernel 
whose shape is controlled by the support set $\Gamma$. The above arguments show that: 
\begin{equation}
{\rm rank}(\mathbf K_{\Gamma}) \leq |\Gamma|-|\Gamma:\Lambda|
\end{equation}
provided the points $\mathbf x_i; i=1,..,N$ lie on the surface \eqref{implicit}.  

If we choose the weighted maps $\phi' = \mathbf D ~\phi$, the kernel function approaches a Gaussian function, periodized to $\left[-1/2,1/2\right]^n$ as $\Lambda \rightarrow \mathbb Z^n$. As $|\Gamma|\rightarrow \infty$, the matrix $\mathbf K_{\Gamma}$ is theoretically full rank. However, we observe that the Fourier series coefficients of a Gaussian function can be safely approximated to be zero outside $|\mathbf k|< 3/\pi\sigma$, which translates to $|\Lambda| \approx \left(\frac{6}{\pi\sigma}\right)^n$; i.e., the rank will be small for high values of $\sigma$. We choose Gaussian kernels since they are more isotropic and less oscillatory than the Dirichlet kernel.

\begin{figure}[t!]
\subfloat[Original \& Bandlimited]{\includegraphics[trim=0 200 0 200,clip,width=0.15\textwidth]{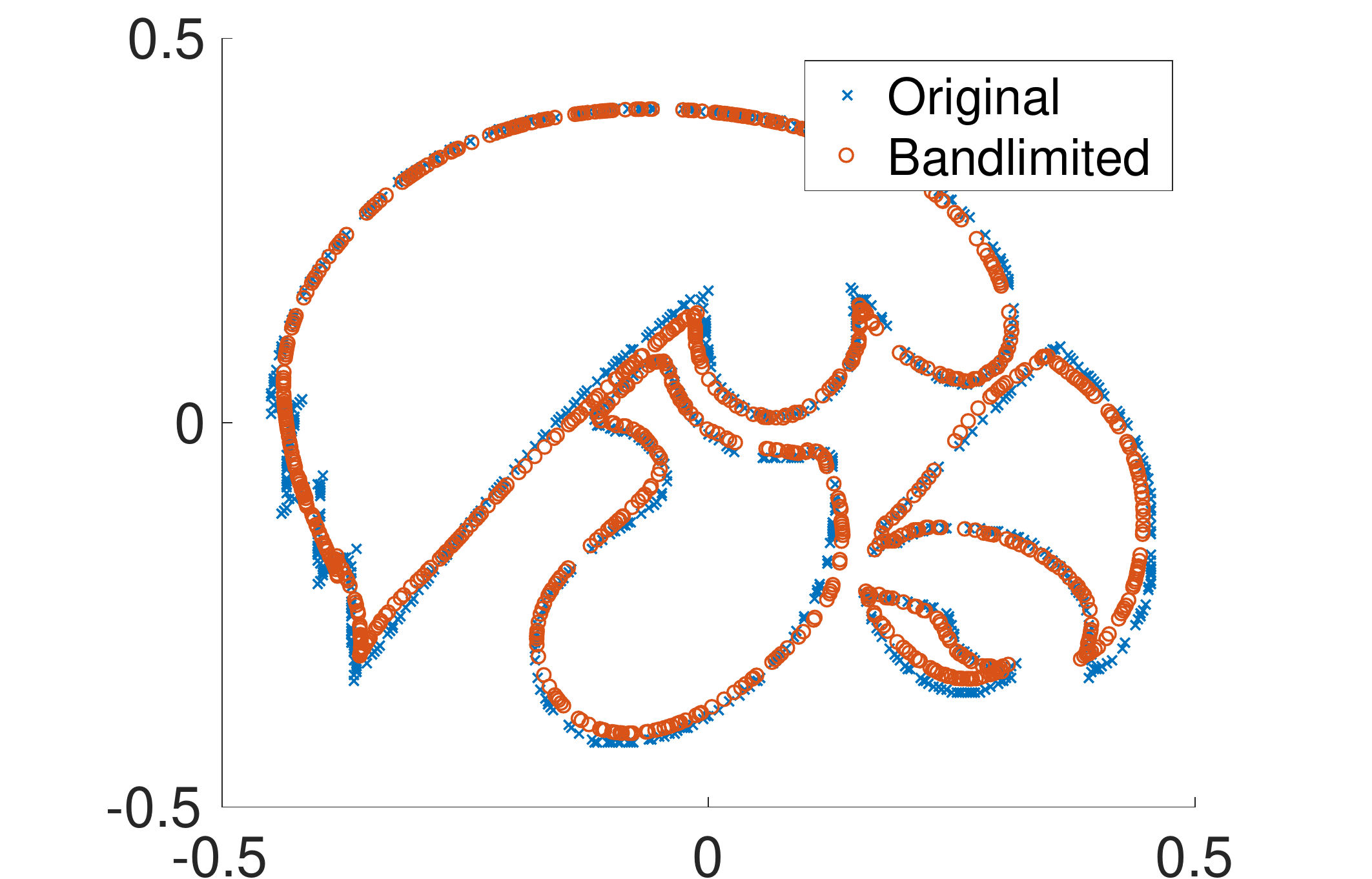}}
\subfloat[Noisy \& Denoised]{\includegraphics[trim=0 200 0 200,clip,width=0.15\textwidth]{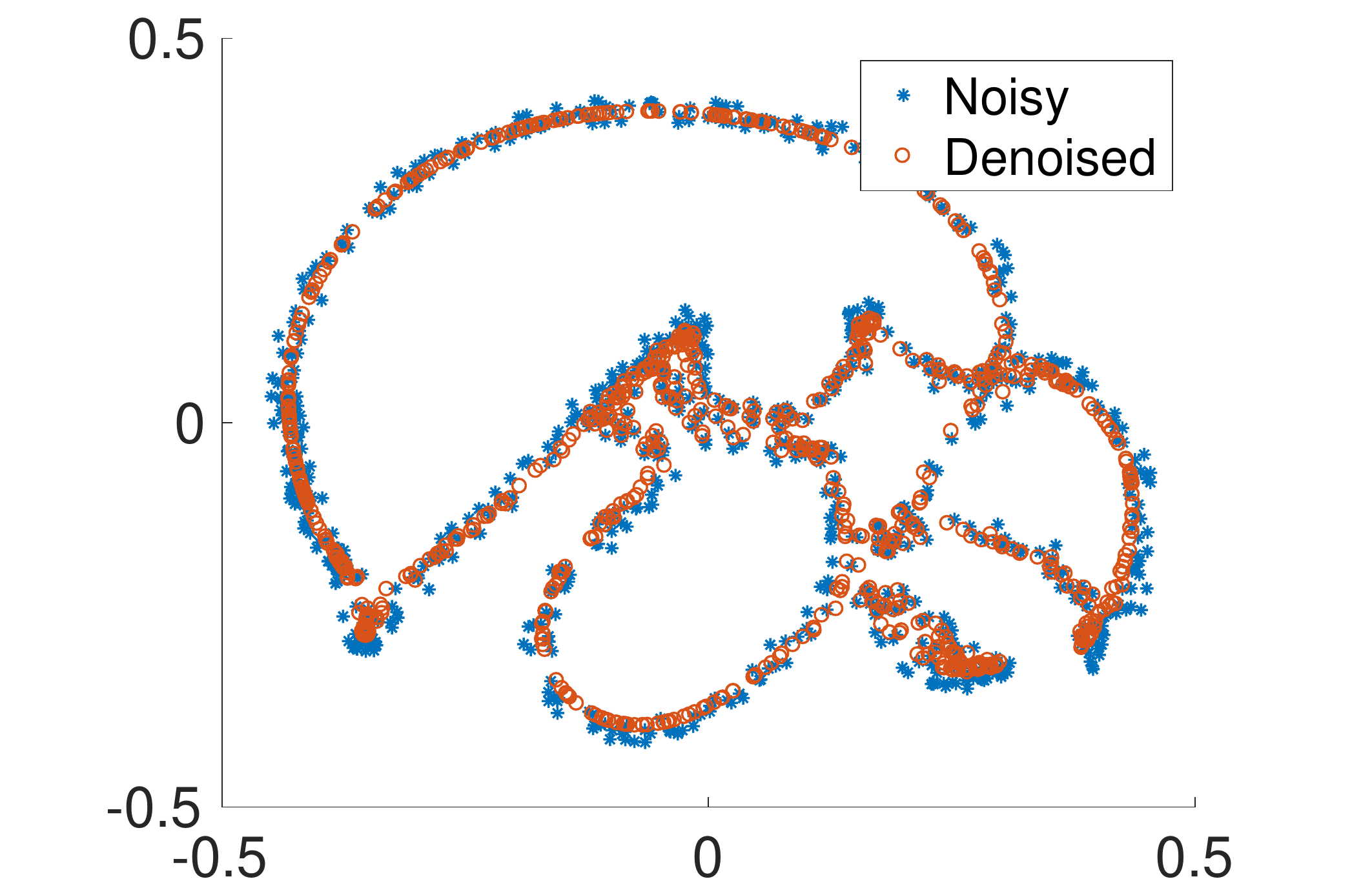}}\vspace{-1em}\\
\subfloat[Singular value decay]{\includegraphics[trim=0 200 0 200,clip,width=0.15\textwidth]{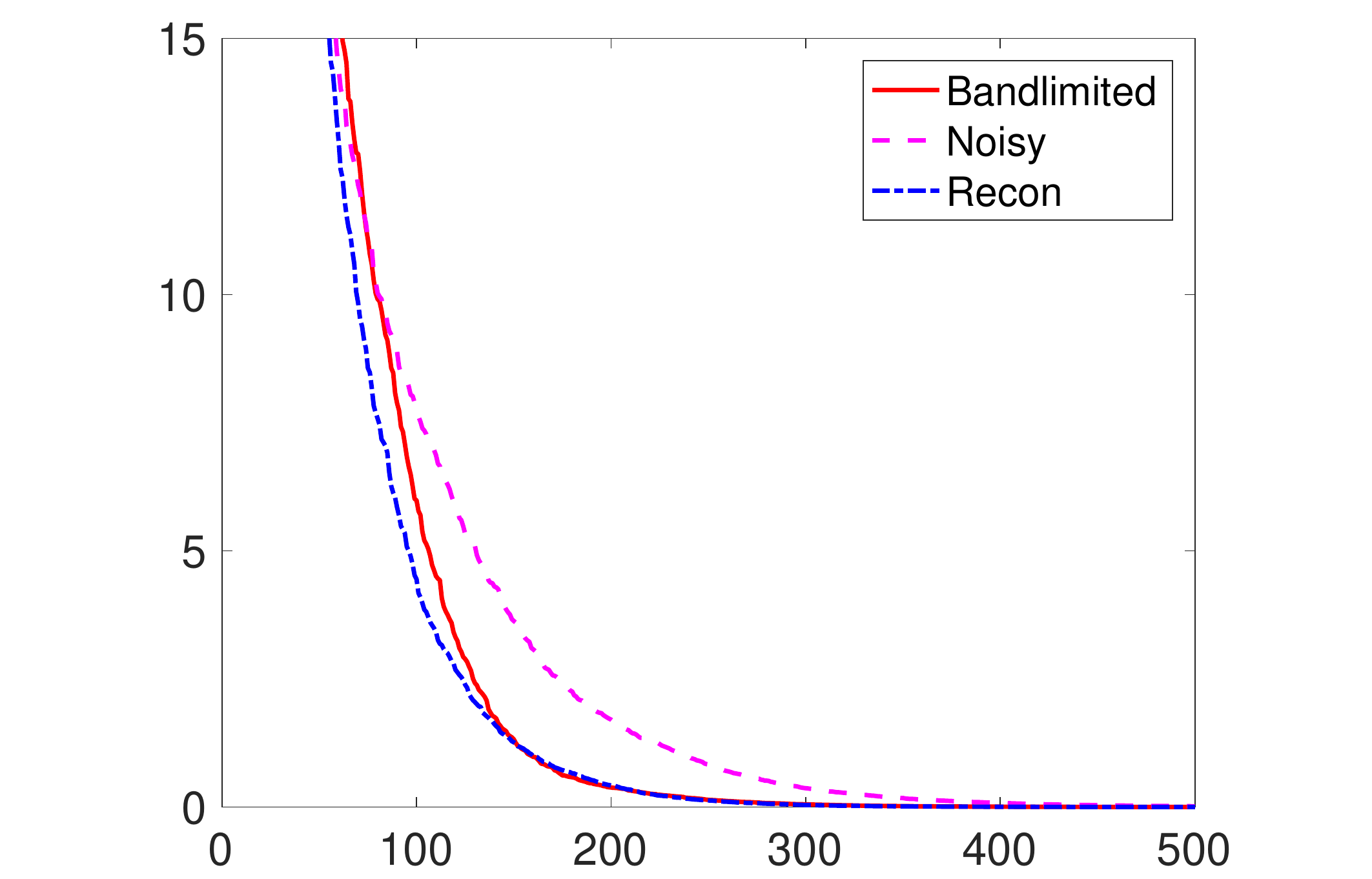}}
\subfloat[Potential function]{\includegraphics[trim=20 300 20 200,clip,width=0.2\textwidth]{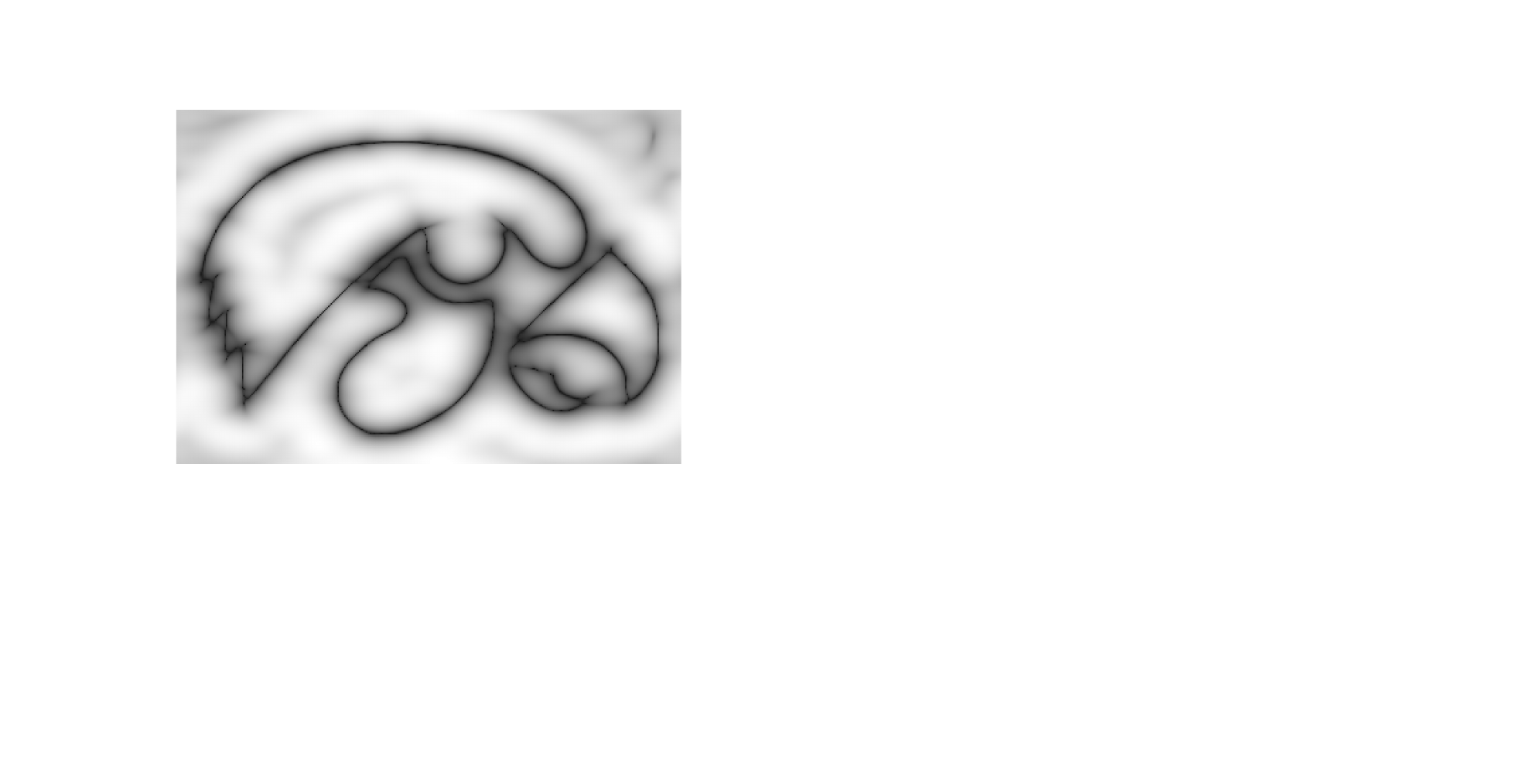}} \vspace{-0.5em}
\caption{Recovery of shape from noisy samples of the Tigerhawk logo. (a) shows the original points, and their projection to a bandlimited Gaussian curve with $\sigma=0.15$. Note from (c) that the singular values of the bandlimited points decay rapidly, while the ones corresponding to the noisy points decay slower. The denoising algorithm \eqref{opt} exploits the lower nuclear norm of $\Phi(\mathbf X)$ to denoise the data. The points after random noise addition and denoising using our algorithm is displayed in (b). Since $\Phi(\mathbf X)$ has several null space vectors, we display the sum of squares function in (d). }\vspace{-1.0em}
\label{Fig1}
\end{figure}

\vspace{-1em}
\subsection{Recovery of noisy point clouds in high dimensions}
We rely on the low rank structure of the kernel matrix $\mathbf K$ to recover the noisy points. Specifically, with the addition of noise, the points deviate from the zero-level set of $\psi$. A high bandwidth potential function is needed to represent the noisy surface, which translates to a high rank matrix $\mathbf K$. Hence, we propose to use the nuclear norm of the feature vectors as a regularizer in the recovery of the cloud from noisy data:
\begin{equation}
\label{opt}
\min_{\mathbf X} \|\mathcal A(\mathbf X) - \mathbf b\|^2 + \lambda\|\mathbf \Phi(\mathbf X)\|_*
\end{equation}

For the case of recovery of points from noisy samples, $\mathcal A(\mathbf X) = \mathbf X$, while in the more general case $\mathcal A$ could be an under-sampling operator. The above formulation is illustrated in Fig. \ref{Fig1}, where we demonstrate the denoising of a shape from its samples; see caption for details. While we illustrate this approach in 2D, it is general enough to be applied in any dimension. However, the direct evaluation of the maps is computationally prohibitive in higher dimensions. We hence now propose an efficient algorithm to denoise the points exploiting the low rank structure of the maps. 
\vspace{-1em}
\subsection{Iterative reweighted least squares (IRLS) algorithm}
We use the IRLS algorithm to solve optimization problem \eqref{opt}. By the definition of the nuclear norm:
\begin{displaymath}
\|\mathbf \Phi(\mathbf X)\|_* = {\rm trace}[(\mathbf \Phi(\mathbf X)^T \mathbf \Phi(\mathbf X))^{\frac{1}{2}}] \approx {\rm trace}[(\mathbf X)\mathbf Q]
\end{displaymath}
where $\mathbf Q = [\mathcal K(\mathbf X) + \gamma \mathbf I]^{-\frac{1}{2}}$. We use this property to solve \eqref{opt} using an alternating strategy:
\begin{equation}
\label{sub1}
\mathbf X^{(n)} = \arg \min_{\mathbf X} \|\mathcal A(\mathbf X) - \mathbf b\|^2 + \lambda~ {\rm trace}[\mathcal K(\mathbf X)\mathbf Q^{(n-1)}]
\end{equation}
where
\begin{equation}
\label{Qeq}
\mathbf Q^{(n)} = [\mathcal K(\mathbf X^{(n)}) + \gamma^{(n)} \mathbf I]^{-\frac{1}{2}}
\end{equation}
Note that the solution for \eqref{sub1} involves a system of non-linear equations. We use gradient linearization to simplify our computations.  Since $\mathcal K(\mathbf X)$ is a Gaussian kernel matrix, linearizing the gradient of the objective function of \eqref{sub1} at each iterate results in the following equivalent optimization problem for \eqref{sub1}:
\begin{equation}
\label{sub1new}
\mathbf X^{(n)} = \arg \min_{\mathbf X} \|\mathcal A(\mathbf X) - \mathbf b\|^2 + \lambda {\rm trace}(\mathbf X^T \mathbf L^{(n-1)} \mathbf X),
\end{equation}
where $\mathbf L^{(n)} = \mathbf D^{(n)} - \mathbf W^{(n)}$. Here, the entries of $\mathbf W^{(n)}$ are $
\mathbf W_{ij}^{(n)} = \frac{f'(\mathbf x_i^{(n)}-\mathbf x_j^{(n)})}{\|\mathbf x_i^{(n)}-\mathbf x_j^{(n)}\|}~~ \mathbf Q_{ij}^{(n)}$
and $\mathbf D^{(n)}_{ii} = \sum_j \mathbf W^{(n)}_{ij}$. The function $f$ determining the weight matrix depends on the Radial Basis Function kernel as: $[\mathcal K(\mathbf X)]_{ij} = f(\|\mathbf x_i-\mathbf x_j\|)$.

We observe the equivalence of the above optimization strategy with widely used non-local means and graph optimization schemes. These schemes estimate a Laplacian matrix $\mathbf L$ or equivalently a weight matrix $\mathbf W$, followed by the minimization of the cost function \eqref{sub1new}. These approaches can thus be seen as fitting a smooth bandlimited surface to the point cloud of patches or signals that are assumed to be on the graph. 

For the denoising problem (where $\mathcal A(\mathbf X) = \mathbf X$), and for some operators $\mathcal A$ it is convenient to solve problem \eqref{sub1new} analytically. In other cases, a conjugate-gradient algorithm can be used to solve it. Note that in the proposed IRLS iterations, it is sufficient to only compute the kernel matrix $\mathcal K (\mathbf X)$; the matrix of feature vectors $\Phi(\mathbf X)$ is never required to be computed explicitly. This is important since the feature vectors in many cases may be large or even infinite dimensional.
\vspace{-1.5em}
\subsection{Application to cardiac MRI}
We apply the proposed framework to the recovery of free breathing and ungated MRI data from highly undersampled measurements. Since MRI is a slow imaging modality, the standard practice in functional cardiac MRI is to integrate the data from multiple heart beats in a breath-held acquisition. However, this approach is often challenging in pediatric and obese subjects who cannot hold their breath. The image frames can be safely assumed to be non-linear functions of two parameters: cardiac and respiratory phase, and hence can be modeled as points on a smooth manifold in high dimensional space. In our previous work, we have acquired the data with navigators, from which the graph Laplacian is estimated. The main challenge with this strategy is the need for customized sequences and the additional 40\% overhead in acquiring the navigators. In this paper we enable a navigator-free acquisition scheme using the proposed approach. 

\vspace{-2em}
\section{Results}
\vspace{-1em}
\begin{figure}[t!]
\centering
\center{\includegraphics[width=0.3\textwidth]{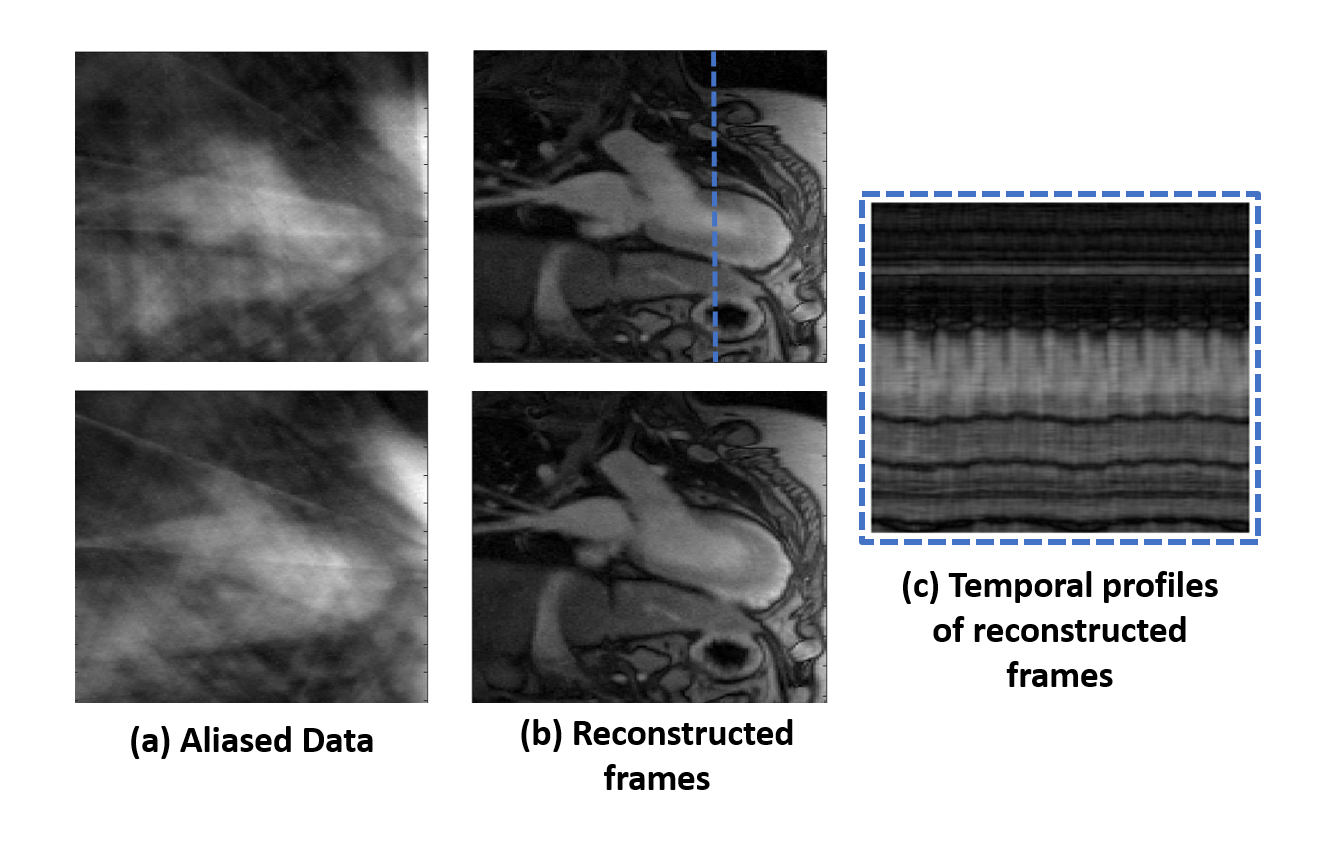}}\vspace{-1.5em}
\caption{Reconstruction of free breathing and ungated cardiac MRI data from $\approx$ 50 fold undersampled measurements. Recovery of the data using standard CS or low-rank methods is challenging due to the extensive motion; see \cite{storm} for comparisons with state of the art methods. (a) denotes two frames, recovered by gridding. (b) Corresponding frames reconstructed using the proposed scheme. (c) Temporal profile from reconstructed frames at the position marked by dotted blue line. The image quality of the proposed scheme is comparable to breath-held methods.}\vspace{-1.5em}
\label{Figcardiac}
\end{figure}

We use the algorithm \eqref{opt} to directly recover the dynamic MRI dataset $\mathbf X$ from golden angle acquired dataset with 10 radial spokes per frame (~50 fold acceleration) and a 30 channel cardiac and spine array. The data was acquired in 40 seconds, which corresponds to 1000 frames and a temporal resolution of $\approx$ 40 ms. A subset of 15 coils was chosen for the reconstruction. The proposed technique was used to iteratively estimate the $\mathbf L$ matrix from the $31 \times 31$ centre under-sampled k-space data. This $\mathbf L$ matrix was used to reconstruct the full data in a single iteration. The in-vivo cardiac free-breathing images recovered from their $\approx 50$ fold undersampled measurements using the proposed scheme are shown in Fig. \ref{Figcardiac}. Two reconstructed frames are shown from the dataset along with the temporal profile. It is observed that the proposed scheme can reconstruct images of good quality while preserving the temporal dynamics.

\vspace{-1em}
\section{Conclusion}
\vspace{-1em}
We introduce a novel framework for the recovery of images/patches, which lie on a smooth low-dimensional manifold/surface in high dimensional space. We model the surface as the level set of a bandlimited function. We show that the non-linear feature maps of the points satisfy annihilation relations. Since these relations imply that the maps lie in a finite dimensional subspace, we use the nuclear norm of the maps as a regularizer to recover the points/images from highly undersampled measurements. The application of this framework to navigator-free free breathing and ungated cardiac MRI provides promising results. 

\vspace{-1.5em}
\bibliographystyle{IEEEbib}
\bibliography{refs}

\end{document}